\documentclass[conference]{IEEEtran}
\IEEEoverridecommandlockouts
\usepackage{cite}
\usepackage{amsmath,amssymb,amsfonts}
\usepackage{algorithmic}
\usepackage{graphicx}
\usepackage{textcomp}
\usepackage{xcolor}
\def\BibTeX{{\rm B\kern-.05em{\sc i\kern-.025em b}\kern-.08em
    T\kern-.1667em\lower.7ex\hbox{E}\kern-.125emX}}

\usepackage[pagebackref=true,breaklinks=true,colorlinks,bookmarks=false]{hyperref}

\usepackage[labelformat=simple]{subcaption}

\usepackage{multirow}
\usepackage{multicol}
\usepackage{xfrac}

\begin{document}

\title{
Query matching for 
spatio-temporal action detection
with query-based object detector
\thanks{This work was supported in part by **** **** **** *****.}
}

\author{
\IEEEauthorblockN{Shimon Hori, Kazuki Omi, Toru Tamaki}
\IEEEauthorblockA{\textit{Nagoya Institute of Technology}, Japan \\
s.hori.110@nitech.jp, toru.tamaki@nitech.ac.jp
}
}

\maketitle

\begin{abstract}
In this paper, we propose a method that extends the query-based object detection model, DETR, to spatio-temporal action detection, which requires maintaining temporal consistency in videos. Our proposed method applies DETR to each frame and uses feature shift to incorporate temporal information. However, DETR's object queries in each frame may correspond to different objects, making a simple feature shift ineffective. To overcome this issue, we propose query matching across different frames, ensuring that queries for the same object are matched and used for the feature shift. Experimental results show that performance on the JHMDB21 dataset improves significantly when query features are shifted using the proposed query matching.
\end{abstract}


\section{Introduction}

In recent years, spatio-temporal action detection (STAD) has received considerable attention \cite{Bhoi_arXiv2019_Spatio-Temporal_Action_Recognition_Survey}, which detects the bounding boxes of human actions in each frame and predicts action categories, resulting in the sequence of bounding boxes. This sequence is referred to as an action \emph{tube} \cite{Gkioxari_2015_CVPR} or a \emph{tubelet} for a video clip.
As with recent progress in object detection \cite{Jiao_IEEEAccess2019_Object_Detection_survey,Carion_ECCV2020_DETR,Redmon_CVPR2016_YOLO,Shaoqing_NIPS2015_Faster-R-CNN,AdaMixer_Gao_CVPR2022,Sparce_RCNN_Sun_CVPR2021}, recent STAD models are built on top of these object detection models, extending their functionality from images to videos \cite{Gritsenko_arXiv2023_STAR,Zhao_CVPR2022_TubeR,Wu_2023_CVPR_STMixer,Gurkirt_WAVC2023_TAAD,Kopuklu_arXiv2019_YOWO,Chen_ICCV2021_WOO,Chang_2023_CVPR_DOAD,Pan_CVPR2021_ACAR-Net}.
However, applying object detection frame by frame does not maintain temporal consistency. Therefore, it is necessary to model temporal \cite{Wu_2023_CVPR_STMixer,Chang_2023_CVPR_DOAD,Pan_CVPR2021_ACAR-Net} or spatio-temporal information \cite{Wu_2023_CVPR_STMixer,STAR_Alexey_2023_arXiv} of videos.

However, the complex architectures of these STAD models have resulted in significant modifications to the original object detection models. This raises the question: How can we adapt original models to STAD with minimal modifications? This study addresses this by extending the query-based object detection model, DETR \cite{Carion_ECCV2020_DETR}, to STAD.
In the proposed method, DETR is applied to each frame, while feature shift \cite{Lin_2019ICCV_TSM} is used to model temporal information. Shifting features to the previous and next frames is a straightforward but effective way to incorporate temporal modeling into frame-by-frame processing. However, in DETR, \emph{object queries}, a representation of features of scene objects, may correspond to different objects in each frame. This makes it difficult to simply shift the queries.
In this paper, we address this issue by proposing query matching between different frames. Our approach ensures that queries for the same object across different frames are matched and used for the feature shift.

\begin{figure}[t]
  \centering
  \includegraphics[width=\linewidth]{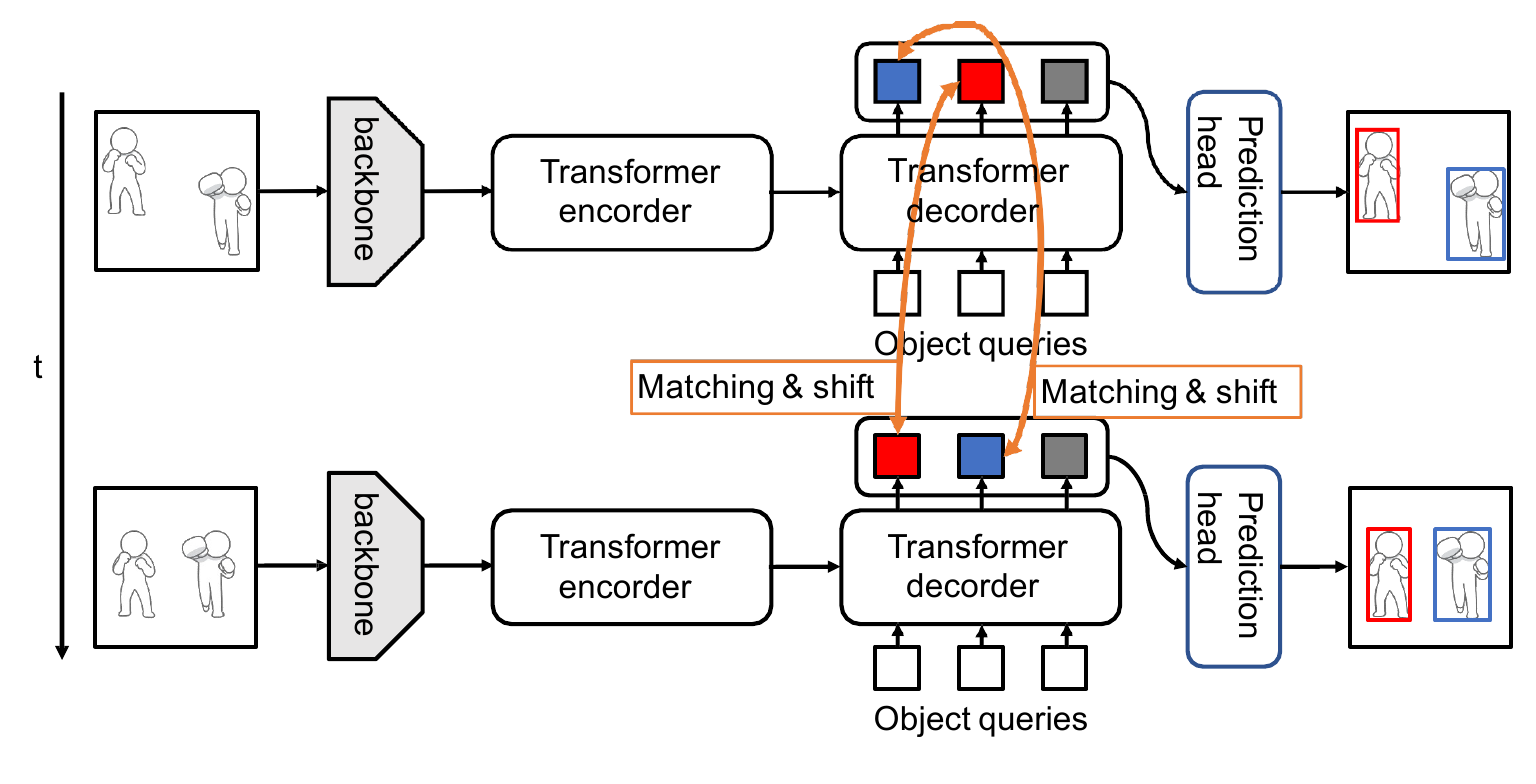}
  \caption{Matching of queries between frames. The positions of queries indicate the same index, while colors indicate the same object.}
  \label{fig:matching}
\end{figure}

\section{Related Work}

Although two-stage and one-stage methods for object detection have been proposed \cite{arkin_survey_2023} in the last decade, methods using object queries, such as DETR \cite{Carion_ECCV2020_DETR} and follow-up work \cite{deformableDETR_arXiv2021_zhu,Meng_ICCV2021_COnditional_DETR,yao_arXiv2021_efficient_DETR,AdaMixer_Gao_CVPR2022}, have recently gained significant attention. DETR learns an object query to each object to be detected and uses Hungarian matching to compute a loss between predicted bounding boxes and ground truth objects.

For STAD, two-stage methods were initially developed \cite{Kopuklu_arXiv2019_YOWO,Chen_ICCV2021_WOO,Sun_ECCV2018_ACRN,Pan_CVPR2021_ACAR-Net}. These methods detect proposals of bounding boxes for each frame, predict their classes, and then post-process by linking bounding boxes across frames to output a final tube. Later, one-stage methods were proposed that directly output tubelets \cite{Gritsenko_arXiv2023_STAR,Zhao_CVPR2022_TubeR,Kalogeiton_ICCV2017_ACT,Wu_2023_CVPR_STMixer} for a given video clip. These methods often use recent object detection models as their foundation.
Of particular relevance to this work are STAR \cite{Gritsenko_arXiv2023_STAR} and TubeR \cite{Zhao_CVPR2022_TubeR}, which extend the DETR's object query to action tubes. However, these models significantly modified the original DETR architecture, keeping only the concept of queries. In this study, our aim is to preserve the DETR's architecture as much as possible while extending it to model temporal information.

For the task of action recognition,
feature shift \cite{Lin_2019ICCV_TSM,Zhang_ACMMM2021_TokenShift,Hashiguchi_2022_ACCVW_MSCA} has been proposed as  an effective method of modeling temporal information without introducing additional computational costs or parameters.
The concept of shift for videos was initially introduced by TSM \cite{Lin_2019ICCV_TSM}, where a 2D CNN is applied to each frame followed by late fusion to produce a prediction of the video level. This approach, while enabling 2D CNN to handle videos, only considers temporal modeling at the final late fusion stage. To address this, TSM incorporated shifting modules within the 2D CNN that shift features forward and backward in time, facilitating temporal interaction between frames. 
This method has been further developed to enable the 2D Vision Transformer \cite{Zhang_ACMMM2021_TokenShift,Hashiguchi_2022_ACCVW_MSCA} to model the temporal information of videos without incurring any additional computational cost or parameters.

However, the feature shift assumes that the channels of features at different times represent the same information of the frame. This assumption is reasonable for action recognition that encodes the whole frame in a single feature tensor, but not for query-based object detection. In this case, DETR's queries in different frames may represent different objects in the scene. Therefore, a simple feature shift might not be effective, and it has not been applied to query-based object detection models.

\section{Shift and matching queries}

In this section, we will first explain the concept of feature shifting, then how this feature shift can be effectively applied to the queries in DETR, by introducing query matching.

\subsection{Temporal feature shift}

Suppose $z_{in} \in \mathbb{R}^{T\times D}$ is an input feature of dimension $D$, which is a stack of frame features of a video clip of $T$ frames.
In the following, $z_{in, t, d}$ denotes the element at $(t, d)$ in $z_{in}$.
It is fed to shifting modules \cite{Hashiguchi_2022_ACCVW_MSCA} to produces output $z_{out}\in \mathbb{R}^{T\times D}$, a temporally shifted version of the input as follows;
\begin{align}
    z_{\mathrm{out}, t, d}
        &= 
    \begin{cases}
        z_{\mathrm{in}, t-1, d}, & \text{$1 < t \le T, 1 \le d < D_f$}\\
        z_{\mathrm{in}, t+1, d}, & \text{$1 \le t < T, D_f \le d < D_f+D_d$}\\
        z_{\mathrm{in}, t, d},   & \text{$\forall t, D_f+D_d \le d \le D$}\\
    \end{cases}
    \label{eq:feature shift}
\end{align}
This indicates that
\begin{itemize}
    \item channels from 1 to $D_f$ are shifted to the next time $t + 1$,
    \item channels from $D_f$ to $D_f + D_d$ are shifted to the previous time $t - 1$, and
    \item the remaining channels from $D_f + D_d$ to $D$ are kept without any shift.
\end{itemize}

Clearly, this feature shift only works when the channel indexes have the same meaning. However, the object queries of DETR may refer to different objects for each frame, so this naive shift cannot be applied.

\subsection{DETR}

DETR \cite{Carion_ECCV2020_DETR} consists of a CNN backbone, a transformer encoder, a transformer decoder, and a prediction head. 
First, the CNN backbone embeds the image and this visual embedding information is fed into the transformer encoder. This encoder output is provided to the transformer decoder as cross-attention. The input to the transformer decoder is the \emph{object query}, which are learnable parameters, and its output represents each object to be detected. These outputs are then fed into the prediction head, where the boxes and classes are predicted.

There are various options for where to insert the feature shift module: within the modules (backbone, encoder, decoder) or at the input or output of these modules. Furthermore, there is flexibility in the positions within the modules for the feature shift, such as inside or outside residual connections, as demonstrated in \cite{Lin_2019ICCV_TSM}.

The object query is the input of the transformer decoder, which is trainable and modified through the decoder with the cross-attention from the output of the encoder. The output of the decoder, which we also call \emph{queries} (shown as colored squares in Fig.\ref{fig:matching}), is fed to the prediction head of a 3-layer MLP that predicts the coordinates and class of the bounding box of each detected object. In this way, the feature shift of the query is considered to be the most effective, since one query holds the information of the bounding box and class of one detected object.

\subsection{Query matching}

Since a query corresponds to one detected object, simply shifting the feature of queries with the same index over frames as with Eq.\eqref{eq:feature shift} could lead to shifting the feature of queries that correspond to different objects in different frames. Therefore, we propose a method to match queries over frames that correspond potentially to the same object, as show in Fig.\ref{fig:matching}.

First, we calculate the cosine similarity for each of the $N$ queries in adjacent frames, resulting in $N \times N$ combinations of similarity in total. We then find the matching of queries by solving a bipartite matching problem using the Hungarian algorithm. Specifically, we find the optimal permutation $\hat{\sigma}$ as follows;
\begin{equation}
    \hat{\sigma} = \underset{\sigma \in \mathfrak{S}} {\operatorname{argmin}}
    \sum_{i} \mathcal{L}_{\mathrm{match}} (q_{i}^{t}, q_{\hat{\sigma}(i)}^{t+1}),
\end{equation}
where $\mathcal{L}_{\mathrm{match}}$ is the cosine similarity
between the $i$-th query $q_{i}^{t}$ at time $t$ and the query $q_{\hat{\sigma}(i)}^{t+1}$ of index $\hat{\sigma}(i)$ at time $t + 1$,
and $\mathfrak{S}$ is a set of permutations.

This query matching is performed on all adjacent frames, and then feature shift is applied to the matched queries.

\section{Experiments}

\subsection{Experimental Setup}

JHMDB21 \cite{Jhuang_ICCV2013_JHMDB21} was used to train and evaluate the proposed method.
It consists of a training set of 660 videos and a validation set of 268 videos,
and each video has up to 48 frames, with annotations for 21 human actions.
In every frame, a single action label and a bounding box are annotated,
resulting in a single action tube for a video, with the same length with the number of frames.

We used DETR \cite{Carion_ECCV2020_DETR}, pre-trained on COCO \cite{Lin_ECCV2014_COCO}, and fine-tuned the prediction head from scratch with the original DETR's loss function with different number of categories.
The loss function we used is the cross-entropy, the same as used in DETR. Note that we used $N=100$ object queries, which is the same as the default of the original DETR, to predict up to $N$ boxes. This number is the upper limit for the predictions, allowing the detection of as many bounding boxes as the number of queries. Therefore, multiple bounding boxes of actions might be predicted for the same person.

\begin{figure}[t]
  \centering

    \begin{minipage}[t]{0.23\linewidth}
      \centering

      \includegraphics[width=\linewidth]{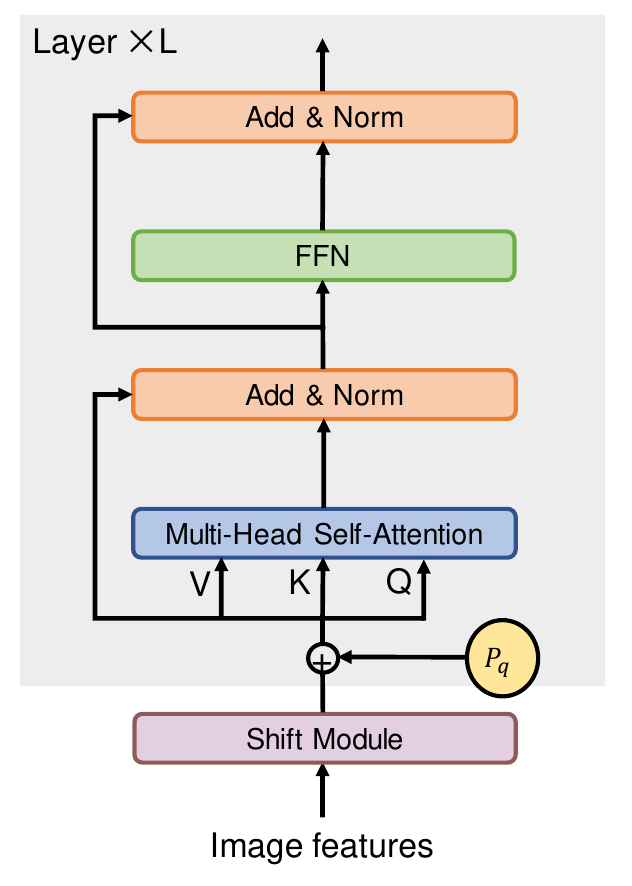}

      \subcaption{}
      \label{fig:encoder_input}
    \end{minipage}
    \hfill
    \begin{minipage}[t]{0.23\linewidth}
      \centering

      \includegraphics[width=\linewidth]{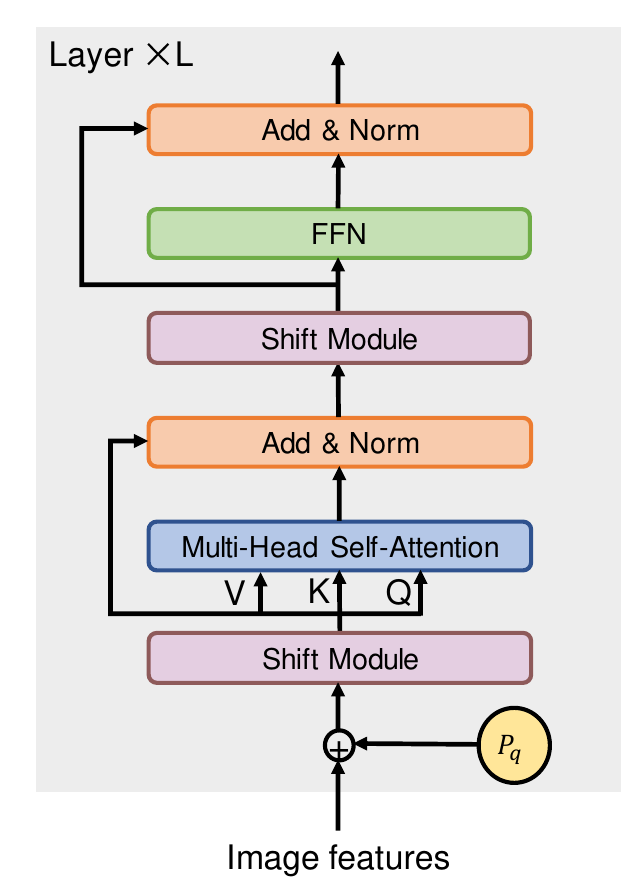}

      \subcaption{}
      \label{fig:encoder_prior_residual}
      
    \end{minipage}
    \hfill
    \begin{minipage}[t]{0.23\linewidth}
      \centering

      \includegraphics[width=\linewidth]{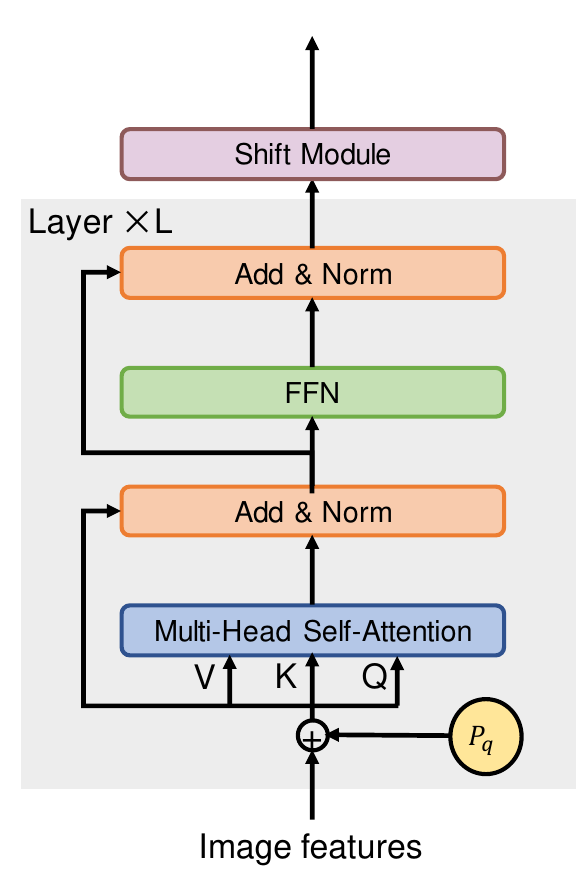}

      \subcaption{}
      \label{fig:encoder_output}
      
    \end{minipage}
    \hfill
    \begin{minipage}[t]{0.25\linewidth}
      \centering

      \includegraphics[width=\linewidth]{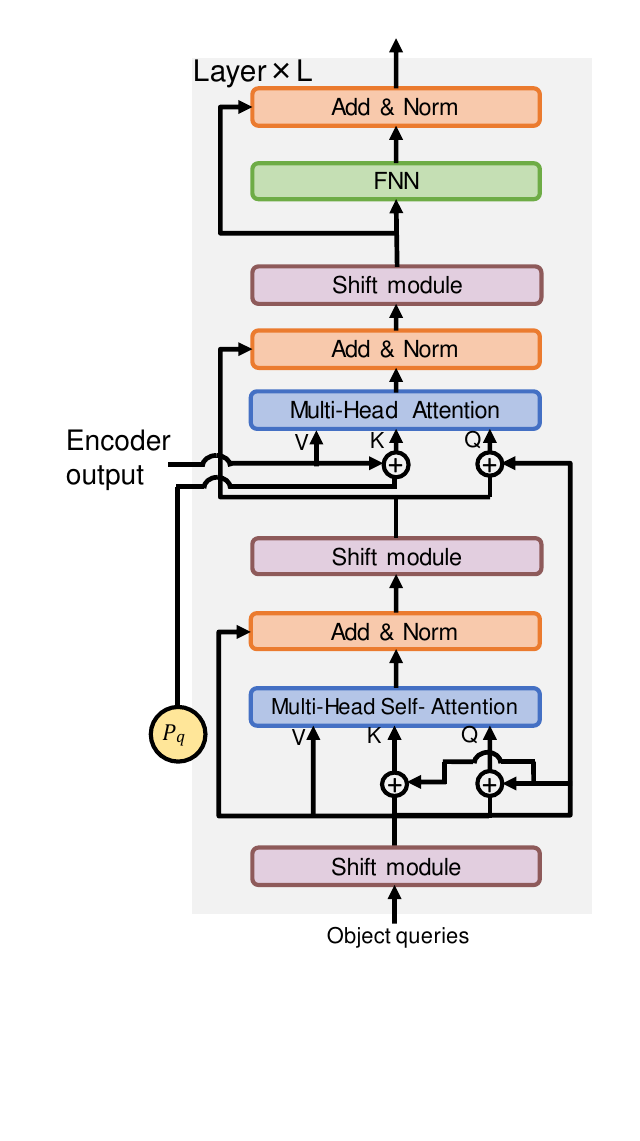}

      \subcaption{}
      \label{fig:decoder_prior_residual}
    \end{minipage}

  \caption{Positions of shit modules;
  (a) encoder input,
  (c) encoder,
  (b) encoder output, and 
  (d) decoder.
  }
  \label{fig:shift locations}

\end{figure}

We extracted an 8-frame video clip with a stride of 1 from a video and fed it into the model. We applied a feature shift at different positions in the model, as shown in Figure \ref{fig:shift locations}.
In each position of shift, we shift $\sfrac{1}{8}$ or $\sfrac{1}{4}$ feature channels out of $D$ channels. As the final prediction of a given video is an action tube, we used the box IoU and the prediction scores to link the predicted bounding boxes across frames.

As evaluation metrics, we used frame-mAP and video-mAP which are commonly used for STAD.
Frame-mAP is mAP based on the bounding box detected for each frame. Predictions are considered correct if the IoU between the predicted and true bounding boxes is greater than a set threshold.
On the other hand, video-mAP evaluates mAP based on the 3D IoU, which measures the overlap between the predicted and the true tubes. 
In the experiments, IoU thresholds of 0.5, 0.75, and an average of every 0.05 from 0.5 to 0.95 (0.5:0.95) were used for evaluation, following a common procedure in previous STAD work. The results were calculated five times each, and the average was used as the final result.

\begin{figure*}[t]
  \centering
    \begin{minipage}[t]{\linewidth}
      \centering
      \includegraphics[width=\linewidth]{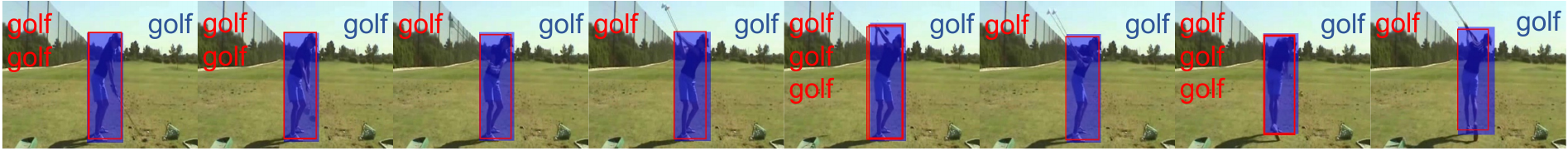}
      \subcaption{}
      \label{fig:shift0_golf}
    \end{minipage}
    \hfill
    \begin{minipage}[t]{\linewidth}
      \centering
      \includegraphics[width=\linewidth]{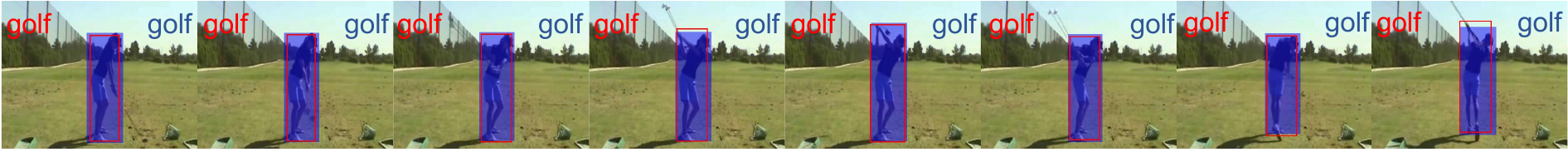}
      \subcaption{}
      \label{fig:query_shift_golf}

      \end{minipage}
    \begin{minipage}[t]{\linewidth}
      \centering

      \includegraphics[width=\linewidth]{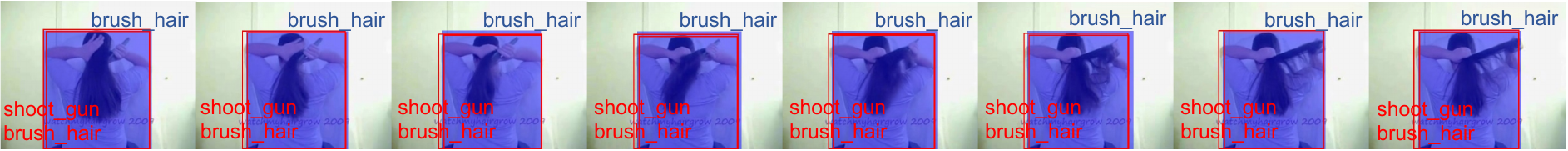}

      \subcaption{}
      \label{fig:shift0_brush_hair}
      
    \end{minipage}
    \begin{minipage}[t]{\linewidth}
      \centering

      \includegraphics[width=\linewidth]{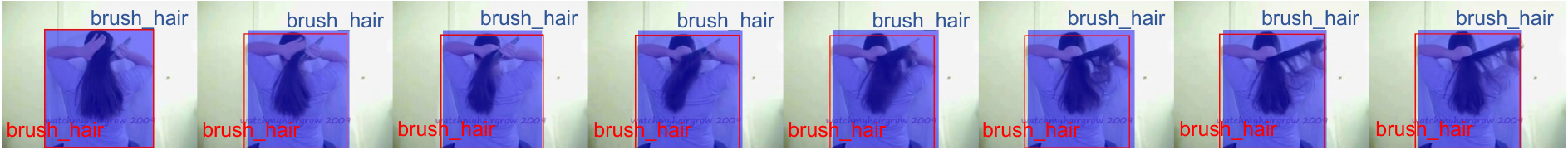}

      \subcaption{}
      \label{fig:query_shift_brush_hair}
    \end{minipage}

    \caption{
    Visualization results of two videos in JHMDB21.
    The blue rectangles represent the ground truth,
    while red rectangles represent detected action tubes,
   (a)(c) without feature shift, and
   (b)(d) with feature shift of $\sfrac{1}{4}$ for queries.
  }
  \label{fig:visualization}

\end{figure*}

\begin{table}[t]

  \caption{The performance comparison with different shift positions.
  Performance differences from the baseline is shown in parentheses.
}

  \label{table:jhmdb}
  \centering

    \begin{tabular}{p{7em}|@{}c@{}|c@{ }c@{ }c@{ }|@{ }c@{ }c@{ }c}
        \multirow{2}{*}{shift position} & 
        \multirow{2}{*}{shift} & 
        \multicolumn{3}{c|}{frame-mAP} & 
        \multicolumn{3}{c}{video-mAP}
        \\
        & & 0.5:0.95&0.5&0.75&0.5:0.95&0.5&0.75 \\
        \hline
        no shift & 0 &24.5&32.6&29.3&11.0&18.9&10.9 \\
        \hline
        \multirow{4}{*}{CNN backbone}
        &$\sfrac{1}{8}$&19.1&26.4&22.7&10.3&17.5&10.5 \\ 
        &              &(-5.4)&(-6.2)&(-6.6)&(-0.7)&(-1.4)&(-0.4) \\ 
        &$\sfrac{1}{4}$&20.6&28.2&24.7&10.6&17.7&11.6\\
        &              &(-3.9)&(-4.4)&(-4.6)&(-0.4)&(-1.2)&(+0.7) \\ 
        \hline
        \multirow{4}{*}{encoder input}
        &$\sfrac{1}{8}$&23.2&31.1&27.7&12.1&19.2&12.5 \\ 
        &              &(-1.3)&(-1.5)&(-1.6)&(+1.1)&(+0.3)&(+1.6) \\ 
        &$\sfrac{1}{4}$&24.5&33.0&29.1&11.6&19.0&12.2 \\
        &              &($\pm$0.0)&(+0.4)&(-0.2)&(+0.6)&(+0.1)&(+1.3) \\ 
        \hline
        \multirow{4}{*}{encoder}
        &$\sfrac{1}{8}$&22.3&30.4&26.4&11.1&17.8&12.1 \\ 
        &              &(-2.2)&(-2.2)&(-2.9)&(+0.1)&(-1.1)&(+1.2) \\ 
        &$\sfrac{1}{4}$&25.4&34.3&29.8&11.2&18.8&11.3\\
        &              &(\textbf{+0.9})&(\textbf{+1.7})&(+0.5)&(+0.2)&(-0.1)&(+0.4) \\ 
        \hline  
        \multirow{4}{*}{encoder output}
        &$\sfrac{1}{8}$&22.3&29.8&26.8&10.5&17.8&10.7 \\ 
        &              &(-2.2)&(-2.8)&(-2.5)&(-0.5)&(-1.1)&(-0.2) \\ 
        &$\sfrac{1}{4}$&24.0&31.9&28.4&12.0&19.4&12.7 \\
        &              &(-0.5)&(-0.7)&(-0.9)&(+1.0)&(+0.5)&(+1.8) \\ 
        \hline  
        \multirow{4}{*}{decoder}
        &$\sfrac{1}{8}$&25.1&33.4&30.4&13.7&21.9&14.8 \\ 
        &              &(+0.6)&(+0.8)&(\textbf{+1.1})&(+2.7)&(+3.0)&(+3.9) \\ 
        &$\sfrac{1}{4}$&22.4&30.1&26.8&14.5&23.0&16.3 \\
        &              &(-2.1)&(-2.5)&(-2.5)&(+3.5)&(+4.1)&(+5.4) \\ 
        \hline  
        \multirow{4}{*}{\parbox{7em}{query\\ \scriptsize{without matching}}}
        &$\sfrac{1}{8}$&22.9&30.7&27.2&12.0&19.1&12.6 \\ 
        &              &(-1.6)&(-1.9)&(-2.1)&(+1.0)&(+0.2)&(+1.7) \\ 
        &$\sfrac{1}{4}$&22.6&30.0&27.1&14.7&22.6&16.3 \\
        &              &(-1.9)&(-2.6)&(-2.2)&(+3.7)&(+3.7)&(+5.4) \\ 
        \hline
        \multirow{4}{*}{\parbox{7em}{query\\ \scriptsize{with matching}}}
        &$\sfrac{1}{8}$&22.8&30.8&27.4&13.6&21.7&14.3 \\ 
        &              &(-1.7)&(-1.8)&(-1.9)&(+2.6)&(+2.8)&(+3.4) \\ 
        &$\sfrac{1}{4}$&24.1&32.5&28.7&15.9&24.7&17.0 \\
        &              &(-0.4)&(-0.1)&(-0.6)&(\textbf{+4.9})&(\textbf{+5.8})&(+6.1) \\ 
        \hline  
        \multirow{4}{*}{\parbox{7em}{query \& decoder\\ \scriptsize{with matching}}}
        &$\tfrac{1}{8} \& \tfrac{1}{8}$ & 24.4&31.5&29.1&15.4&23.7&17.3\\
        &              &(-0.1)&(-1.1)&(-0.2)&(+4.4)&(+4.8)&(\textbf{+6.4}) \\ 
        &$\tfrac{1}{4} \& \tfrac{1}{4}$&22.5&29.9&26.6&13.9&22.0&16.0 \\
        &             &(-2.0)&(-2.7)&(-2.7)&(+2.9)&(+3.1)&(+5.1) \\ 
        &$\tfrac{1}{8} \& \tfrac{1}{4}$&23.9&31.5&28.4&14.2&22.2&15.7 \\
        &              &(-0.6)&(-1.1)&(-0.9)&(+3.2)&(+3.3)&(+4.8) \\ 
        &$\tfrac{1}{4} \& \tfrac{1}{8}$&24.0&32.1&28.6&14.3&22.3&15.5 \\
        &             &(-0.5)&(-0.5)&(-0.7)&(+3.3)&(+3.4)&(+4.6) \\ 
        \end{tabular}
    
\end{table}

\subsection{Quantitative evaluation}

Table \ref{table:jhmdb} shows the performance of frame-mAP and video-mAP for different shift positions.
The baseline performance without any shift is shown in the top row, and the performance differences from the baseline are shown in parentheses.

When CNN backbone features were shifted, frame-mAP decreased by approximately 6\%, although video-mAP did not show a significant performance deterioration. Other studies using feature shift \cite{Lin_2019ICCV_TSM,Zhang_ACMMM2021_TokenShift,Hashiguchi_2022_ACCVW_MSCA} also reported only marginal performance changes, suggesting that the effect of feature shift within the CNN backbone was also ineffective in the proposed method.

Applying feature shift in the encoder (input, inside, or output) slightly improved frame-mAP, only by up to 2\%. However, video-mAP, a more critical metric for STAD, showed no significant differences, or even deteriorated.
The reason might be that modifications of features by the shifts do not directly impact the output of the prediction head.

Feature shift in the decoder resulted in an increase in video-mAP by 3\% to 5\%. This suggests that using feature shift in the decoder is effective because of the direct impact on the output.
When the queries (i.e., decoder output) were shifted, the performance of video-mAP significantly improved compared to the shift in the decoder. However, without the proposed query matching, performance is not improved. In addition, frame-mAP performance decreases substantially when query matching is not utilized. This highlights the effectiveness of the proposed query matching.

Based on the observation above, we applied feature shift to both the decoder and queries. Despite the video-mAP increasing by as much as 6.4\%, the decrease in frame-mAP was minimal.
Improving both the frame-mAP and video-mAP is difficult. However, as mentioned above, given the significance of the video-mAP in this STAD task, it seems that shifting both the query and the decoder is most effective.

\subsection{Quantitative evaluation}

Figure \ref{fig:visualization} shows the visualization of the detected tubes by sifting queries with query matching. Without feature shifting, multiple classes were detected for a single ground truth, or the same class was detected more than once. On the other hand, with feature shifting in object queries, there was a tendency for one correct class to be detected for each ground truth. Without feature shifting, over-detection was observed, suggesting that over-detection was suppressed by shifting features in the object query.

\section{Conclusion}

In this paper, we proposed an STAD method that extends DETR through feature shift and query matching. Experimental results demonstrated that shifting the decoder features and queries with the proposed query matching outperforms cases where matching is not used and when shifting modules are integrated into the backbone and encoder. For future research, we plan to evaluate our method on AVA \cite{Gu_2018CVPR_AVA-Actions}, a larger and more complex dataset for STAD, investigate why feature shifts in queries and decoders improved performance, and compare performance with the prior work.

\section*{Acknowledgment}
This work was supported in part by JSPS KAKENHI Grant Number JP22K12090.

\bibliography{mybib,all}
\bibliographystyle{ieeetr}

\end{document}